\documentclass{article}
\usepackage{arxiv}
\usepackage{enumitem}
\usepackage{notation}
\usepackage{tikz}
\usetikzlibrary{bayesnet}
\usepackage{subcaption}
\usepackage[textwidth=80pt]{todonotes}

\usepackage{graphics}
\usepackage{tikz}
\usepackage{titlesec}
\usepackage{setspace}
\usepackage[colorlinks=true, linkcolor=red, urlcolor=blue, citecolor=blue, anchorcolor=blue]{hyperref}
\usepackage{cleveref}
\usepackage[authoryear,round]{natbib}

\usetikzlibrary{shapes,decorations,arrows,calc,arrows.meta,fit,positioning}
\tikzstyle{node_fill}=[circle,draw,preaction={fill=black!20},font={}]
\tikzset{
    -Latex,auto,node distance =0.7 cm and 0.7 cm,semithick,
    state/.style ={circle, draw, minimum width = 0.7 cm},
    point/.style = {circle, draw, inner sep=0.04cm,fill,node contents={}},
    bidirected/.style={Latex-Latex,dashed},
    el/.style = {inner sep=2pt, align=left, sloped}
}
\crefname{figure}{Fig.}{Figs.}
\crefname{definition}{Defn.}{Defns.}
\crefname{corollary}{Corollary}{Corollaries}
\crefname{proposition}{Prop.}{Props.}
\crefname{theorem}{Thm.}{Thms.}
\crefname{remark}{Remark}{Remarks}
\crefname{principle}{Principle}{Principles}
\crefname{lemma}{Lemma}{Lemmas}
\crefname{claim}{Claim}{Claims}
\crefname{table}{Tab.}{Tabs.}
\crefname{section}{\S}{\S\S}
\crefname{subsection}{\S}{\S\S}
\crefname{subsubsection}{\S}{\S\S}

\title{Algorithmic Recourse in Partially and Fully Confounded Settings Through Bounding Counterfactual Effects}

\author{
Julius von K\"ugelgen\thanks{Corresponding author; \texttt{jvk@tuebingen.mpg.de}} \,$^{1,2}$ 
\quad Nikita Agarwal $^{3}$
\quad Jakob Zeitler $^4$
\quad Afsaneh Mastouri $^4$
\quad Bernhard Sch\"olkopf $^1$\\\\
$^{1}$ Max Planck Institute for Intelligent Systems T\"ubingen, Germany\\
$^2$ Department of Engineering, University of Cambridge, United Kingdom\\
$^3$ Graduate Training Centre of Neuroscience,
International Max Planck Research School,\\ University of T\"ubingen, Germany\\
$^4$ University College London, United Kingdom\\
}

\begin{document}

\maketitle

\begin{abstract}
Algorithmic recourse aims to provide actionable recommendations to individuals to obtain a more favourable outcome from an automated decision-making system.
As it involves reasoning about interventions performed in the physical world, recourse is fundamentally a \textit{causal} problem. 
Existing methods compute the effect of recourse actions using a causal model learnt from data under the assumption of no hidden confounding and modelling assumptions such as additive noise.
Building on the seminal work of~\citet{balke1994counterfactual}, 
we propose an alternative approach for \textit{discrete} random variables which relaxes these %
assumptions and allows for unobserved confounding and arbitrary structural equations.
The proposed approach only requires specification of the causal graph and confounding structure and 
bounds the expected counterfactual effect of recourse actions. 
If the lower bound is above a certain threshold, i.e., 
on the other side of the decision boundary,
recourse is guaranteed in expectation.
\end{abstract}

\section{Introduction}
Black-box machine learning (ML) models are increasingly used for consequential decision-making, e.g., to predict credit or recidivism risk based on an individual's features~\citep{chouldechova2017fair}.
While a growing literature aims to provide explanations \textit{why} a particular prediction was made~\citep{wachter2017counterfactual}, granting agency to individuals dictates that they should, in principle, be able to obtain a more favourable prediction by \textit{actively improving their situation}~\citep{venkatasubramanian2020philosophical}.
Algorithmic recourse aims to automate the process of providing individuals with actionable recommendations to remedy their situation~\citep{ustun2019actionable,karimi2020survey}.

Since actions carried out in the real world may have downstream effects on some variables but not on others, reasoning about such hypothetical interventions, as in the context of algorithmic recourse, is fundamentally a causal problem~\citep{karimi2020mint}. 
It thus requires \textit{causal assumptions} about the data generating process, i.e., the underlying (socio-economic) system.
A common assumption is that the causal graph of the observed variables is known from expert knowledge and domain understanding.
To compute the causal effect of recourse actions, however, this is insufficient on its own: additional assumptions such as the absence of unobserved confounders, and/or modelling assumptions such as linearity or additive noise are needed.
Existing approaches rely on such assumptions to learn a causal model from data that can be used to reason about the effect of recourse actions~\citep{karimi2020imperfect}.
Since these are strong assumptions which are typically violated in real-world settings,
we argue that such reliance decreases the credibility of the drawn conclusions.

We therefore propose a new approach for algorithmic recourse which assumes that only the causal graph and the observational distribution of features are known, i.e., we do not assume a particular parametric form of the structural equations and allow for unobserved confounding.
Our approach requires all observed variables to be discrete and is based on the computation of bounds on causal queries, also known as partial identification, applied to recourse.
We adapt existing methodology with provably tight bounds to algorithmic recourse with full confounding, and introduce a new formulation for the partially confounded case.

\subsection{Related work}
Bounding of causal effects was first extensively discussed by~\citet{manski1990nonparametric}. \citet{balke1994counterfactual}~then introduced bounding in structural causal models (SCMs), based on a reformulation of the SCM with response function variables. %
While most work has focused on discrete variables and specific graphs (such as instrumental variable models), recent work attempts to generalise these ideas to continuous variables~\citep{kilbertus2020class,zhang2021bounding} and arbitrary graphs~\citep{sachs2020symbolic,noam2020class,bareinboim2021nonparam,hu2021generative}. 
While~\citet{wu2019pc} have applied causal bounds for algorithmic fairness, we are not aware of existing work in the context of algorithmic recourse.
For a more detailed review, we refer to~\citet{richardson2014nonparametric}.

\section{Problem setting}

Let $\Xb=(X_1, ..., X_n)$ denote 
random variables, or
\textit{features},
(e.g., age, occupation, income, etc) taking values in $\Xcal=\Xcal_1\times ...\times \Xcal_n$, and let $h:\Xcal\rightarrow [0,1]$ be a given (probabilistic) classifier that was trained to predict a binary decision variable (e.g., whether a loan was approved or denied).
For an individual, or factual observation, $\xF=(x_1^\Ftt, ..., x_n^\Ftt)$ that obtained an unfavourable classification, $h(\xF)<0.5$, algorithmic recourse aims to answer 
what they could have done, or could do, to flip the decision~\citep{ustun2019actionable}.
\looseness-1 Since this question involves reasoning about changes, or interventions, carried out in the physical world, addressing it requires a causal description of the data generating process.

\paragraph{Causal model.}
We adopt the 
framework of~\citet{pearl2009causality} and assume that the generative process is
governed by an (unknown) structural causal model (SCM)
$\Mcal=(\fb, \PP_\Ub)$, i.e., each 
$X_i$ is generated according to a structural equation 
\begin{equation}
    \label{eq:original_SCM}
    X_i:=f_i(\PA_i, U_i), \quad \quad \text{for} \quad \quad i=1, ..., n,
\end{equation}
where $\PA_i\subseteq \Xb\setminus X_i$ are the causal parents, or direct causes, of $X_i$; $f_i$ are deterministic functions; and $\Ub=(U_1, ..., U_n)$ are unobserved exogenous random variables with unknown joint distribution $\PP_{\Ub}$.
\textit{Crucially, we do not assume that~$\PP_\Ub$ factorises, thus allowing for unobserved confounding.}
The causal graph~$\Gcal$ associated with~$\Mcal$---obtained by drawing an edge from each variable in $\PA_i$ to $X_i$ for all $i$, thus summarising the qualitative causal relations between features---is assumed acyclic and known, see~\cref{fig:problem_setting} for an example.

\newcommand{\xshift}{2em}
\newcommand{\yshift}{2em}
\begin{figure}
\centering
\begin{subfigure}[b]{0.33\columnwidth}
    \centering
    \begin{tikzpicture}
    \centering
    \node (X_1) [latent] {$X_1$};
    \node (X_2) [latent, below=of X_1, xshift=-\xshift, yshift=\yshift] {$X_2$};
    \node (X_3) [latent, below=of X_1, xshift=\xshift, yshift=\yshift] {$X_3$};
    \edge{X_1}{X_2,X_3};
    \edge{X_2}{X_3};
    \end{tikzpicture}
    \caption{unconfounded}
    \label{fig:unconfounded}
\end{subfigure}%
\begin{subfigure}[b]{0.33\columnwidth}
    \centering
    \begin{tikzpicture}
    \centering
    \node (X_1) [latent] {$X_1$};
    \node (X_2) [latent, below=of X_1, xshift=-\xshift, yshift=\yshift] {$X_2$};
    \node (X_3) [latent, below=of X_1, xshift=\xshift, yshift=\yshift] {$X_3$};
    \edge{X_1}{X_2,X_3};
    \edge{X_2}{X_3};
    \path[bidirected] (X_1) edge[bend right=60] (X_2);
    \path[bidirected] (X_2) edge[bend right=60] (X_3);
    \end{tikzpicture}
    \caption{partially conf.}
    \label{fig:partially_confounded}
\end{subfigure}%
\begin{subfigure}[b]{0.33\columnwidth}
    \centering
    \begin{tikzpicture}
    \centering
    \node (X_1) [latent] {$X_1$};
    \node (X_2) [latent, below=of X_1, xshift=-\xshift, yshift=\yshift] {$X_2$};
    \node (X_3) [latent, below=of X_1, xshift=\xshift, yshift=\yshift] {$X_3$};
    \edge{X_1}{X_2,X_3};
    \edge{X_2}{X_3};
    \path[bidirected] (X_1) edge[bend right=60] (X_2);
    \path[bidirected] (X_2) edge[bend right=60] (X_3);
    \path[bidirected] (X_1) edge[bend left=60] (X_3);
    \end{tikzpicture}
    \caption{fully confounded}
    \label{fig:fully_confounded}
\end{subfigure}
\caption{Overview of different assumptions: dashed bi-directed arrows indicate confounding, i.e., the existence of an unobserved common cause, as manifested by a dependence between the corresponding exogenous variables $U_i$ (not shown). Existing work on causal recourse assumes no hidden confounding as in (a), whereas the present work addresses the confounded settings (b) and (c).}
\label{fig:problem_setting}
\end{figure}
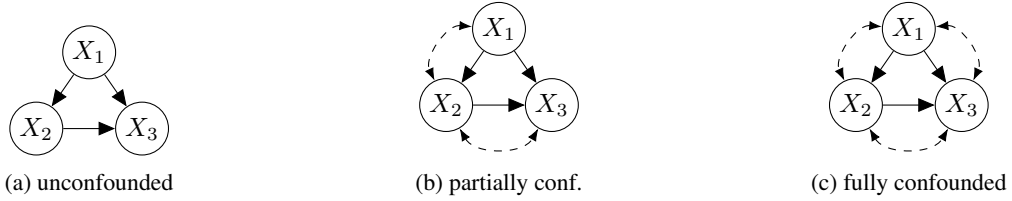

\paragraph{Recourse optimisation problem.}
Given a causal model,
\citet{karimi2020mint}
propose to 
address the algorithmic recourse problem
for individual $\xF$ by \textit{finding a set of minimal interventions that would have led to a changed prediction}, i.e., by solving the following optimisation problem,
\begin{equation}
\label{eq:recourse_optimisation_problem}
    \min_{\thetaI \in \Fcal(\xF)}  \text{ cost}(\thetaI;\xF) 
    \,\,\,\, \text{s.t.} \,\,\,\, h\left(\xb_{do(\thetaI)}(\ub^\Ftt)\right)>0.5,%
\end{equation}%
where $\Fcal(\cdot;\xF)$ is a set of feasible interventions $do(\thetaI)$ which assign the value $\thetaI$ to a subset of variables $\Xb_\Ical\subseteq \Xb$ with $\Ical\subseteq\{1,..,n\}$;
$\text{cost}(\thetaI;\xF)$ is a cost function measuring the effort required of $\xF$ for $do(\thetaI)$; 
and $\xb_{\thetaI}(\ub^\Ftt)$ denotes the \textit{structural counterfactual}, or counterfactual twin, of $\xF$ that would have occurred according to $\Mcal$ if $do(\thetaI)$ had been performed, all else being equal.
It is computed from $\Mcal$ by fixing the exogenous variables $\Ub$ to their factual value $\ub^\Ftt$ (\textit{abduction}), replacing the structural equations for $\Xb_\Ical$ by $\Xb_\Ical:=\thetaI$ (\textit{action}), and computing the effect on the descendants $\Xb_\dI$ of $\Xb_\Ical$ (\textit{prediction}).

\paragraph{Assumptions for recourse.}
Computing counterfactual queries 
requires full SCM specification~\citep{pearl2009causality,peters2017elements},
but the underlying SCM $\Mcal$ is typically unknown.
Even if the SCM is fully specified, it is not always possible to uniquely infer the factual value $\uF$ of $\Ub$ corresponding to individual $\xF$.
In practice, the counterfactual query $h\left(\xb_{do(\thetaI)}(\ub^\Ftt)\right)$ in~\eqref{eq:recourse_optimisation_problem} therefore needs to be replaced with the expected classification w.r.t.\ the counterfactual distribution under $do(\thetaI)$ given $\xF$, that is
\begin{equation}
\label{eq:probabilistic_recourse_constraint}
    \EE_{\Ub|\xF}\left[ h\left(\xb_{do(\thetaI)}(\Ub)\right)\right]\,.
\end{equation}

Existing methods then aim to solve a probabilistic version of~\eqref{eq:recourse_optimisation_problem} with a constraint based on~\eqref{eq:probabilistic_recourse_constraint} by using a (family of) approximate SCMs $\widehat{\Mcal}$ which can be learnt from data under strong additional assumptions (besides a known causal graph), such as no hidden confounding (i.e., fully-factorised $\PP_\Ub$, see~\cref{fig:unconfounded}) and structural constraints on the $f_i$ in~\eqref{eq:original_SCM} such as additive (Gaussian) noise~\citep{karimi2020imperfect}.\footnote{If a  point estimate of $\Mcal$, learnt under an additive noise assumption, is used, this leads to point-estimate of $\uF$ and thus of the counterfactual~\citep{mahajan2019preserving, karimi2020imperfect}.}
However, such assumptions are often too strong to be realistic: hidden confounding is commonplace in real-world settings, and additive noise only applies to continuous variables and does not allow for heteroscedasticity or multi-modality.
We relax these assumptions and introduce
an approach for causal algorithmic recourse in the presence of unobserved confounding~(see~\cref{fig:partially_confounded,fig:fully_confounded}) and arbitrary structural equations, based on bounds.

\section{Bounding causal effects for recourse}
Counterfactuals are generally not identifiable in the presence of hidden confounding%
~\citep{pearl2009causality}.
We therefore adopt the approach of~\citet{balke1994counterfactual} to
bound%
~\eqref{eq:probabilistic_recourse_constraint} for a given individual~$\xF$ and recourse action~$do(\thetaI)$.
To this end, we require the following additional assumptions:
\begin{enumerate}[label=(\roman*)]
    \item $\forall i: \Xcal_i=\{0, 1, ..., K_i-1\}$, that is, all $X_i$ are discrete random variables with $|\Xcal_i|=K_i$ states each;
    \item the observational distribution $\PP_\Xb$ is known (or can be estimated accurately from data).
\end{enumerate}
The general idea is to first use assumption (i) to reformulate the SCM in a way that allows to parametrise the unknown distribution over exogenous variables $\PP_\Ub$, and then use assumption (ii) to optimise~\eqref{eq:probabilistic_recourse_constraint}
over all distributions which are consistent with the observed $\PP_\Xb$ and the assumed confounding structure.

\begin{figure}[t]
    \centering
    \includegraphics[width=0.75\textwidth]{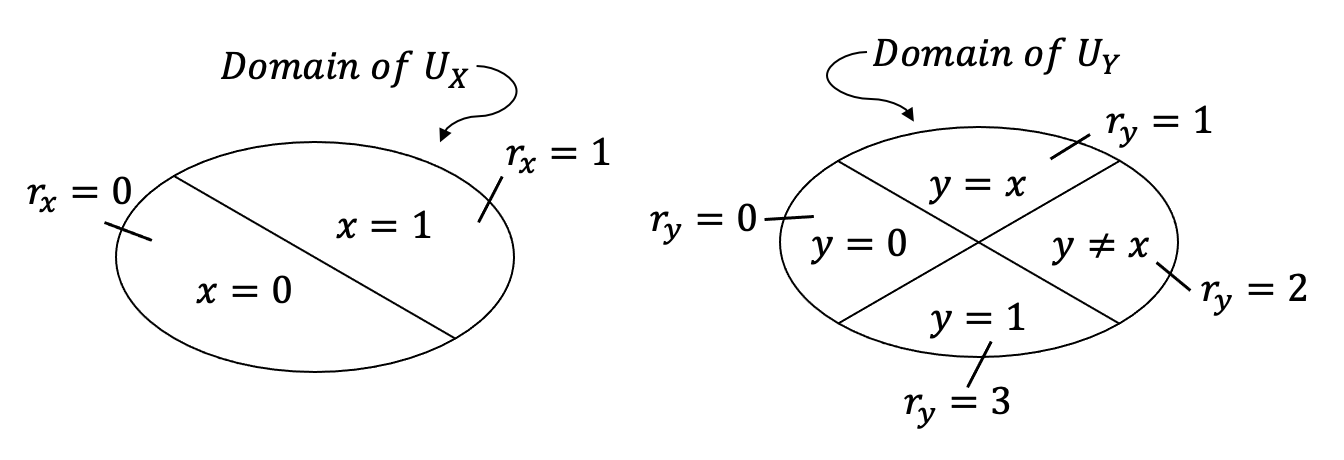}
    \caption{Response function framework for two binary variables $X\rightarrow Y$: only four distinct functions exist from $\Xcal$ to $\Ycal$, so  $\Ucal_Y$ can be split accordingly; the discrete response function variable $R_Y$ then indicates which of the four regions $U_Y$ falls into.}
    \label{fig:response_functions_intuition}
\end{figure}

\subsection{Response-function reformulation}
\label{sec:response_function_reformulation}
Since the domains $\Ucal_i$ of the exogenous $U_i$ are unknown (they could, e.g., be continuous or high-dimensional), we cannot directly parametrise $\PP_\Ub$.
However, since
all $X_i$ are assumed discrete, we can reformulate the SCM $\Mcal$ into an equivalent one where the $U_i$ are replaced by \textit{discrete response function variables} $R_i$~\citep{balke1994counterfactual}.
Intuitively, there are only finitely many distinct functions $m_i$ that map one discrete domain
(that of $\PA_i$) to another (that of $X_i$), and we can think of the $U_i$ as (randomly) determining which function is applied.
We can therefore partition each $\Ucal_i$ into finitely many regions corresponding to these functions, and define a new discrete random variable $R_i$ which indicates which region $U_i$ falls into (i.e., which response function $m_i$ is applied),
see~\cref{fig:response_functions_intuition} for an illustration.

Formally, we replace the original SCM~$\Mcal=(\fb,\PP_\Ub)$ from~\eqref{eq:original_SCM} with an equivalent one~$\Mcal^{\Rb}=(\mb,\PP_\Rb)$
where the response function variables 
$\Rb=(R_1, ..., R_n)$ with
$R_i=l_i(U_i)$ index all possible response functions $m_i(\cdot,R_i)$ for each $i$, that is, we rewrite~\eqref{eq:original_SCM} as%
\begin{equation}
\begin{aligned}
    \label{eq:reparameterized_SCM}
    X_i:=f_i(\PA_i,U_i) 
    =f_i(\PA_i,l_{i}^{-1}(R_i))
    =f_i\circ l_{i}^{-1}(\PA_i,R_i)
    =m_i(\PA_i,R_i),
    \end{aligned}
\end{equation}
see~\cite{balke1994counterfactual} for further details.
The number of response functions, i.e., the size of the domain $\Rcal_i$ for each $R_i$ is 
$|\Rcal_i|=K_i^{\prod_{X_j\in \PA_i}K_j}$ if $\PA_i$  is not empty, and $|\Rcal_i|=K_i$ otherwise; we write $\Rcal=\Rcal_1\times...\times \Rcal_n$.

The advantages of reformulation~\eqref{eq:reparameterized_SCM} are twofold: first, unlike for $\Mcal=(\fb,\PP_\Ub)$, the structural equations $\mb$ of~$\Mcal^\Rb$ are known; and second, while unknown, $\PP_\Rb$ is a discrete distribution which we can easily parametrise.
Any choice of $\PP_\Rb$ leads to a fully specified SCM that induces a unique observational distribution and allows for computing counterfactual queries.
We can thus minimise (maximise) the counterfactual query~\eqref{eq:probabilistic_recourse_constraint}
over all $\PP_\Rb$ which are consistent with the observed $\PP_\Xb$ to obtain a lower (upper) bound. 

\subsection{Bounds for the fully confounded case}
\label{sec:fully_confounded_case}
First, we consider the fully confounded case (see~\cref{fig:fully_confounded}) following the treatment of~\citet{balke1994counterfactual}, adapted to the context of probabilistic causal recourse~\eqref{eq:recourse_optimisation_problem}-\eqref{eq:probabilistic_recourse_constraint}.
Since, in the fully confounded case, $\PP_\Ub$ and hence also $\PP_\Rb$ do not admit a non-trivial factorisation, we directly parametrise the unknown joint distribution as $q_\rb=\PP_\Rb(\Rb=\rb)$.
For notational convenience, we also write $p_\xb=\PP_\Xb(\Xb=\xb)$, and denote by~$\pb$ and~$\qb$ the probability vectors obtained by stacking $p_\xb$ and $q_\rb$ for all $\rb\in\Rcal,\xb\in\Xcal$, respectively.

\paragraph{Constraints.}
The constraint that $\Mcal^\Rb=(\mb,\PP_\Rb)$ needs to be consistent with the observed $\PP_\Xb$, i.e., that any $\PP_\Rb$ needs to be such that it induces $\PP_\Xb$ via $\mb$, can be written as
\begin{equation}
\label{eq:constraint_matching_marginals}
\forall \xb\in\Xcal: \quad  p_{\xb}=\sum_{\rb\in\Rcal} q_{\rb}\prod_{i=1}^n\II\{x_i=m_i(\pa_i,r_i)\}\,.
\end{equation}

Intuitively, the RHS of~\eqref{eq:constraint_matching_marginals} aggregates the probability of all values $\rb$ of $\Rb$ which give rise to given $\xb$; since, generally, $|\Rcal|>|\Xcal|$, there may be multiple such terms.
Collecting the products of indicator functions $\II\{\cdot\}$ on the RHS of~\eqref{eq:constraint_matching_marginals} in a binary $|\Xcal|\times |\Rcal|$ matrix $A$, we obtain $\pb=A\qb$.
Moreover, we have the simplex constraint~$\qb\in\Delta^{|\Rcal|-1}$.

\paragraph{Objective.}
Next, we write the objective to be optimised, i.e., the counterfactual query in~\eqref{eq:probabilistic_recourse_constraint}, for a particular choice of $\PP_\Rb$.
Note that a counterfactual change to $\Xb_\Ical$ will not affect the non-descendant variables $\Xb_{\ndI}$ which will remain fixed at their factual value $\xF_\ndI$.
For any given $do(\thetaI)$, we thus only need to reason about changes to the descendant variables $\Xb_\dI$.
If the set of descendants is empty, we can directly evaluate the classifier and there is no need for bounding.
We thus assume that $\dI$ is not empty.
The query~\eqref{eq:probabilistic_recourse_constraint} can then be written in terms of~$\qb$ as

\begin{equation}
\label{eq:objective_FC}
    \begin{aligned}
    \Lcal(\qb;\xF,\thetaI)
    &=\sum_{\xb_\dI} h\left(\xF_\ndI, \thetaI, \xb_\dI\right)
    \PP\left(\Xb_{\dI; do(\thetaI)}=\xb_\dI|\Xb=\xF\right)
   \\
    &=\frac{1}{p_{\xF}}\sum_{\xb_\dI} h\left(\xF_\ndI, \thetaI, \xb_\dI\right)
    \PP\left(\Xb_{\dI; do(\thetaI)}=\xb_\dI,\Xb=\xF\right)
  \\
    &=
    \frac{1}{p_{\xF}}
    \sum_{\xb_\dI}
    h\left(\xF_\ndI, \thetaI, \xb_\dI\right)
    \sum_{\rb\in\Rcal}
    q_\rb
    \left(
    \prod_{i=1}^n
    \II\left\{x_i^{\texttt{F}}=m_i\left(\pa_i^{\texttt{F}}, r_i\right)\right\}
    \right)
    \left(
    \prod_{i\in\dI}
    \II\left\{x_i=m_i(\pa_{i;do(\thetaI)}, r_i)\right\}
    \right)
    \end{aligned}
\end{equation}

where 
$\pa_i^{\texttt{F}}$ and $\pa_{i;do(\thetaI)}$ denote the 
factual and counterfactual (post-intervention) values of $\PA_i$, respectively.

\paragraph{Optimisation.}
To bound the expected outcome~\eqref{eq:probabilistic_recourse_constraint}
of a particular action $do(\thetaI)$ for a given individual $\xF$, we can then solve
the following optimisation problem:%
\begin{equation}
    \label{eq:optimisation}
        \minmax_{\qb\in\Delta^{|\Rcal|-1}}  \quad \Lcal(\qb;\xF,\thetaI) 
        \quad \text{subject to} \quad \pb=A\qb\,.
\end{equation}

Since both objective~\eqref{eq:objective_FC} and constraint~\eqref{eq:constraint_matching_marginals} are linear in~$\qb$ and the simplex is convex,~\eqref{eq:optimisation} is a linear program which can be solved exactly~\citep{schrijver1998theory} and efficiently  by linear solvers such as \texttt{cvxpy}~\citep{diamond2016cvxpy}.
\vspace{-0.5em}
\subsection{Bounds for the partially confounded case}
\label{sec:partial_confounding}
\vspace{-0.25em}
In~\cref{sec:fully_confounded_case}, we have directly parametrised the joint distribution $\PP_\Rb$, corresponding to the most general case of arbitrary confounding, as shown in~\cref{fig:fully_confounded}.
However, we may know (e.g., from domain experts) that only a subset of the causal relations are confounded.
This case is illustrated in~\cref{fig:partially_confounded} where $X_1$ \& $X_2$ and $X_2$ \& $X_3$ are confounded, but $X_1$ \& $X_3$ are not.
Such knowledge can be useful if available as it further constrains the problem and may, in principle, lead to tighter, and thus more informative, bounds.
We therefore propose a new formulation for this scenario.

We assume that the partial confounding structure is known. 
It implies a non-trivial factorisation of~$\PP_\Ub$, and thus of~$\PP_\Rb$.
For example,~\cref{fig:partially_confounded} implies 
$\PP_\Rb=\PP_{R_1} \PP_{R_2|R_1}  \PP_{R_3| R_2}$.
The last term does not depend on $R_1$ since $X_1$ and $X_3$ are unconfounded.
This allows to parametrise $\PP_\Rb$ with fewer free parameters than in the fully-confounded case from~\cref{sec:fully_confounded_case}.
In general, we assume that $\PP_\Rb$ factorises as
\begin{equation}
\label{eq:partial_confounding_response_factorisation}
\textstyle
    \PP_\Rb=\prod_{i=1}^n \PP_{R_i|\PA(R_i)}
\end{equation}
where $\PA(R_i)\subseteq \Rb_{1:(i-1)}$ 
indicates which 
$X_1, ..., X_{i-1}$ are confounded with $X_i$.\footnote{Note that $\PA_i\neq\PA(R_i)$: the former are the observed parents of $X_i$, the latter the unobserved ``parents'' of $R_i$; e.g., setting $ \forall i: \PA(R_i)=\Rb_{1:(i-1)}$ results in a fully-confounded scenario, while setting $\forall i: \PA(R_i)=\varnothing$ leads to an unconfounded one. }
Instead of the joint distribution, we parametrise each conditional on the RHS of~\eqref{eq:partial_confounding_response_factorisation} as:%
\begin{equation*}
\label{eq:def_si}
\textstyle
    s_{r_i, \pa(R_i)}=\PP\big(R_i=r_i|\PA(R_i)=\pa(R_i)\big).
\end{equation*}%
We then proceed as in the fully confounded case by replacing $q_\rb$ in the set of constraints~\eqref{eq:constraint_matching_marginals} and the objective~\eqref{eq:objective_FC} by
\begin{equation*}
\textstyle
    q_\rb=\PP_\Rb(\Rb=\rb)=\prod_{i=1}^n s_{r_i, \pa(R_i)}.
\end{equation*}%
This leads to a similar optimisation problem to~\eqref{eq:optimisation}, but where we instead optimise over a smaller number of parameters $s_{r_i,\pa(R_i)}$ specifying valid conditional distributions.
However, since the $s_{r_i,\pa(R_i)}$ appear in the form of products (in both the objective and constraints), the resulting optimisation problem is non-convex and thus more challenging to solve.
We discuss possible solutions in~\cref{sec:discussion}.

\vspace{-0.5em}
\subsection{Using bounds to inform recourse}
\vspace{-0.25em}
For a given $\xF$ and $do(\thetaI)$, the proposed approaches from~\cref{sec:fully_confounded_case} and~\cref{sec:partial_confounding} assuming either full (FC) or partial (PC) confounding, respectively, will result in lower (LB) and upper (UB) bounds on~\eqref{eq:probabilistic_recourse_constraint} that satisfy:
\begin{equation*}
    \textsc{lb}_\textsc{fc}
    \leq 
    \textsc{lb}_\textsc{pc}
    \leq
    \EE_{\Ub|\xF}\left[ h\left(\xb_{do(\thetaI)}(\Ub)\right)\right]
    \leq
    \textsc{ub}_\textsc{pc}
    \leq
    \textsc{ub}_\textsc{fc}
\end{equation*}
where the first and last inequality should generally be strict (i.e., the PC bounds tighter) if knowledge about a non-trivial partial confounding structure is available.
Since the aim of recourse is to find actions $do(\thetaI)$ that would result in a changed prediction ($h>0.5$), we are mainly interested in the lower bounds.
In line with~\eqref{eq:recourse_optimisation_problem}, we could, for example, choose to recommend the lowest cost action for which the expected outcome is guaranteed to be larger than $0.5$ (or $0.5+\epsilon$ to be more conservative).
Alternatively, we could present $\xF$ with the bounds and costs for each action to enable a more informed decision.
Finally, we note that here we chose to bound the expected prediction w.r.t.\ the unknown (and unidentifiable) counterfactual distribution, meaning that recourse is only guaranteed \textit{on average} if $\text{LB}>0.5$.
To be more conservative, we could also bound the worst case outcome by replacing the sum over all possible $\xb_\ndI$ in~\eqref{eq:objective_FC} with a minimum instead

\section{Discussion}
\label{sec:discussion}
\paragraph{Discrete variables.}
The bounding approach to recourse proposed in the present work requires discrete variables for the response function reformulation and parametrisation of the unobserved distribution.
If continuous variables are present, these can either be discretised, or treated separately, either via additional structural assumptions~\citep{karimi2020imperfect} or by extending recent advances in bounding for continuous outcomes~\citep{kilbertus2020class}.

\paragraph{Computational constraints.}
The number of response functions grows very quickly for densely connected graphs (i.e., larger $|\PA_i|$) and  with the number of states of the observed variables (i.e., larger $|\Xcal_i|$). Moreover, complex confounding structures require more parameters to specify $\PP_\Rb$.
Computational scalability is a fundamental challenge for computing bounds, which makes this approach currently only feasible for few observed variables with few states.

\paragraph{Optimisation approaches for partial confounding.}
A promising approach to solve the non-convex optimisation problem resulting from the partially confounded case~(\cref{sec:partial_confounding}) appears to be a reformulation in which pairs of $s_{r_i, \pa(R_i)}$ are defined as auxiliary variables, leading to a mixed integer quadratic program which can be solved with bilinear solvers using a branch and bound algorithm as provided, e.g., in \texttt{gurobi}~\citep{gurobi2018gurobi}.
A study of optimality guarantees is relevant for future work.

\paragraph{Preliminary experimental evidence.}
Preliminary experimental evidence suggests that the proposed bounding approach can be useful in that---for some combinations of $h$, $\PP_\Xb$, and $\xF$, and under unobserved confounding---actions are found for which the lower bound on the expected outcome is larger than 0.5.
Moreover, for the partially confounded case, the obtained bounds are typically tighter than those based on assuming full confounding.
We leave a more thorough empirical investigation for future work.

\section{Conclusion}
We proposed the first approach for causal algorithmic recourse in the presence of unobserved confounding.
While counterfactuals are unidentifiable in this case, the expected outcome of recourse actions can be bounded subject to constraints from the observational distribution, provided that all observed variables are discrete.
In the fully confounded case, this leads to a well known formulation as a linear program which can be solved exactly.
When domain knowledge about a partial confounding structure is available, we propose a new formulation that takes this information into account to obtain tighter bounds.
Since the resulting optimisation problem is non-convex, it remains an open question how best to tackle it.
More efficient algorithmic solutions and potential applications to fairness~\citep{gupta2019equalizing,von2020fairness} constitute interesting directions for future work.

\section*{Acknowledgements}
We are grateful to Amir-Hossein Karimi, Chris Russell, Elias Bareinboim, Jia-Jie Zhu, and Ricardo Silva for helpful discussions and comments. 
This work was supported by the German Federal Ministry of Education and Research (BMBF): T\"ubingen AI Center, FKZ: 01IS18039B; and by the Machine Learning Cluster of Excellence, EXC number 2064/1 - Project number 390727645.

\bibliographystyle{plainnat}
\bibliography{references}

\end{document}